\documentclass [10pt,letter]{article}
\usepackage{palatino,epsfig,latexsym,amsmath}
\include{graphics}
\usepackage{algorithm}
\usepackage{algorithmic}
\usepackage{pstricks}
\usepackage{graphicx}
\usepackage{pstricks,pst-plot,pst-node}

\begin {document}
\null
\medskip
\bigskip
\bigskip
\centerline{\LARGE Effective linkage learning using }
\centerline{\LARGE low-order statistics and clustering\nonumber\footnote{Submitted to IEEE Transactions on Evolutionary Computation}}
\vspace{0.5 cm}
\bigskip
\centerline{Leonardo~Emmendorfer$^\dag$, Aurora~Pozo$^\ast$}

\normalsize
\vspace{0.7 cm}
\centerline{$^\dag$Numerical Methods in Egnineering, PhD program}
\centerline{$^\ast$Department of Computer Science }
\centerline{Federal University of Paran\'{a}, Brazil }
\centerline{\{leonardo,aurora\}@inf.ufpr.br}

\vspace{0.5 cm}


\begin{quote}
\noindent{\large \bf Abstract}\\
\small \it
The adoption of probabilistic models for the best individuals found so far is a powerful approach for evolutionary computation. Increasingly more complex models have been used by estimation of distribution algorithms (EDAs), which often result better effectiveness on finding the global optima for hard optimization problems. Supervised and unsupervised learning of Bayesian networks are very effective options, since those models are able to capture interactions of high order among the variables of a problem. Diversity preservation, through niching techniques, has also shown to be very important to allow the identification of the problem structure as much as for keeping several global optima. Recently, clustering was evaluated as an effective niching technique for EDAs, but the performance of simpler low-order EDAs was not shown to be much improved by clustering, except for some simple multimodal problems. This work proposes and evaluates a combination operator guided by a measure from information theory which allows a clustered low-order EDA to effectively solve a comprehensive range of benchmark optimization problems.

\end{quote}

%

\setcounter{page}{1}
\pagestyle{plain}



\line(1,0){394}
\section{Introduction}

Evolutionary algorithms solve optimization problems by evolving successive populations of solutions until convergence occurs. Two steps are usually present at each generation: selection of promising solutions and creation of new solutions in order to obtain a new population. 

Combination of genetic information is a major concern in evolutionary computation. In the simple genetic algorithm (sGA) \cite{hol75} this mechanism is implemented as the crossover operator, which creates a new individual from two parents by combining portions of both strings. Recently, estimation of distribution algorithms (EDAs) started a novel approach for learning information from the best individuals, which involves inferring a probabilistic model and sampling from this model in order to generate the next population. Combination of information is achieved in EDAs since a single model is built from several good individuals. Unfortunately, combining different individuals may lead to poor results if the model adopted is not expressive enough. 

Simpler order-1 EDAs adopt probabilistic models which assume independence among genes. This class of EDA is known for its simplicity and computational efficiency in model learning, since no search for model structure needs to be performed \cite{cga}. Further, the simple conception and implementation should make those algorithms very attractive. Their low effectiveness on harder benchmark problems, however, is unacceptable. This is a major drawback, since genetic and evolutionary algorithms are known for their wide applicability and robustness \cite{pena2005}. 

High order EDAs, on the other hand, are based on learning the linkage among genes by inferring expressive probabilistic models based on searching for a factorization, which captures the dependencies among genes. Good results are reported for several problems in the literature whereas a high computational cost associated to the model induction stage is imposed in this class of EDAs. Finding a factorization can be a computationally expensive process and the resulting graph is often a suboptimal solution \cite{sporadic_model}\cite{yuan_gallagher}.

One of the most important efforts to make EDAs more effective is to adopt clustering as a strong niching approach, inducing the preservation of diversity in the population. Niching is crutial for evolutionary computation in general \cite{mahfoud} and for EDAs in particular \cite{ahn2006}. It improves the identification of the problem structure as much as enhances the chances of finding a higher number of global optima on multimodal problems. K-means clustering algorithm \cite{kmeans} has recently been applied as a niching technique based on grouping genotypically similar solutions together.  The performance of simpler low-order EDAs, however, was not show to be much improved by clustering except for some simple unstructured multimodal problems. Low-order clustered EDAs have not been able to solve hard deceptive structured problems \cite{pelikan2000}.


The main contribution of this paper is to show that a simple low-order EDA aided by clustering the population and guided by information measures is able to perform linkage learning and, therefore, solve a representative set of benchmark problems. This work extends a previous paper \cite{EP07a}, where some of the ideas and results reported here were first presented. Here the foundations of the algorithm and operators proposed are discussed in a more detailed fashion, and a wider set of experiments is presented.

The main difference between the new approach and other similar clustered low-order EDAs is on how to deal with the combination of information. Clustering, or similar unsupervised learning approaches, are usually applied to the population of EDAs in order to prevent combination of different niches (clusters). Conversely, the new operator trusts in the combination of information from different clusters as an effective method for exploring the search space. This combination, however, must be carefully performed, respecting the information carried out by each cluster. 
The new operator captures relevant information about the problem structure from the probabilistic models of the clusters and builds a combined probabilistic model from two different parent clusters, attempting to maintain the most informative part of each parent intact, hence preserving the problem structure.

Experiments illustrate the effectiveness of this combination operator on detecting the problem structure. When the operator is turned off or replaced by a random recombination, the algorithm is not able to find the global optima of any benchmark problem tested. The problems adopted here illustrate several aspects which have been recently considered as tricky for EDAs, such as deception \cite{pelik99}, symmetry \cite{pelikan2000}, hierarchy \cite{hiff}, global multimodality \cite{pena2005} and the presence of overlapping building blocks \cite{yu2005}. The structured fashion of those and other classes of problems makes them hard for low-order and even for high-order EDAs.

This work is organized as follows. Section 2 discusses the main concepts related to the field of estimation of distribution algorithms. Section 3 reviews the most closely related work concerning to the application of clustering in estimation of distribution algorithms. A discussion about how a generic clustered order 1 EDA could detect higher order of interactions is presented in Section 4, where a new operator guided by information measures is proposed.

Section 5 presents the algorithm $\varphi$-PBIL, which implements the ideas discussed in Section 4. A default parameter setting for the algorithm is discussed in Section 6. Section 7 describes the benchmark optimization problems which were adopted in this paper. Section 8 presents empirical verifications of the performance of $\varphi$-PBIL, comparing it to state-of-the-art compentent EDAs. Finally, Section 9 discusses results and implications of this work.

\section{Linkage Learning and Estimation of Distribution Algorithms}

Interacting variables are more informative together than alone. 
Interaction among variables is an irreducible whole; a dependency which cannot be broken \cite{jakulin_bratko}. A classic example from machine learning literature related to the importance of interaction among variables is the XOR problem \cite{minsky} which is not linearly separable and cannot be solved without capturing interactions.

In genetic and evolutionary computation, the identification and preservation of important interactions among genes is called linkage learning and this topic gained increasing attention from the community. Most linkage learning techniques aim at the identification of substructures which should be conserved during combination \cite{harik}. A similar concept in genomics, called genetic linkage, is defined as the association of genes on the same chromosome. When two genes are independent, the Mendelian law of independent assortment states that the segregation of one gene is independent of the segregation of the other \cite{liu}.

Simple genetic algorithm (sGA) with one-point crossover relies on the ordering of the genes in the codification of the problem. In order to achieve success it is required that interacting variables should be coded as nearby genes in the chromosome in such a way that crossover would less probably cause the disruption of substructures. More recently other approaches adopted alternative schemes ranging from the simple reordering of the genes to more complex mechanisms like subspecification and superspecification of solutions, as in Messy GA \cite{messyga}. This algorithm adopts a two-stage evolutionary process where substructures are identified in a first stage and subsequently combined. This ensures that substructures were all correctly identified before going on and exploring the combinations among them.

A different situation occurs in Estimation of Distribution Algorithms (EDAs), where probabilistic modeling is applied as the learning mechanism for the evolutionary process. Algorithm \ref{alg_eda} is the general description of an EDA.

\begin{algorithm}[h]
\caption{A general estimation of distribution algorithm}
\label{alg_eda}

\begin{algorithmic}

\STATE Generate an initial random population.\\
\STATE Compute the fitness of the individuals. \\
\WHILE{convergence criteria are not met}
\STATE Build a probabilistic model from the best individuals in the population.\\
\STATE Sample from this model in order to breed new individuals and update the population.
\ENDWHILE

\end{algorithmic}
\end{algorithm}

An EDA solves a problem by building probabilistic models of the best solutions from the population. New solutions (individuals) are sampled from the model, so the traditional operators of crossover and mutation, which operate at the level of the individuals, are not suited for EDAs. 

The most effective EDAs adopt fairly complex models which should capture the whole structure of dependence among the genes of the problem. The major motivation for learning complex and powerful models is to acquire and maintain information about interactions among the variables of the problem. 
Some earlier EDAs, however, learn probabilistic models which assume independence among all genes, as

\begin{equation} \label{order1} {\pi}( \boldsymbol{x} )=\prod_{i=1}^m {\pi}_i(x_i)\end{equation} \\
\begin{tabular}[ht]{rl}
where & $\pi$ is the joint probabilistic model\\
      &$\boldsymbol{x}$ is the binary vector of an individual\\
      & ${\pi}_i(x_i)$ is the  marginal model of gene i\\
      &\\
\end{tabular}

For a binary codification, the marginal model for each gene is the binomial proportion of 1s in that gene for the selected individuals. Algorithms based on this kind of model are not able to detect any interactions among the genes without additional support, therefore they present a similar behavior as a GA with uniform crossover \cite{cga}. Since no interactions are considered and only order 1 statistics are used, those algorithms should be called order 1 EDAs. One of the most important member of this class is PBIL (Population Based Incremental Learning) \cite{pbil}.  

In PBIL the population is represented by a probability vector  $\boldsymbol{p}=(p_1,v_p,\ldots,p_m)$, as in (\ref{order1}), where $p_j$ represents the probability of an individual to possess a 1 in gene $j$. At each generation, M individuals are generated and the best N are selected to update the model.

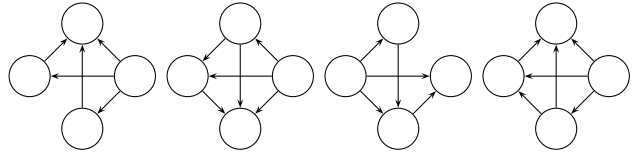
\begin{figure}[t]

\setlength{\unitlength}{0.035cm}
\psset{unit=0.7cm}

\begin{center}
\begin{pspicture}(0,0)(13,4)
\psset{arrows=->,linewidth=0.02}

\cnode(2,3){0.4}{A}
\cnode(3,2){0.4}{B}
\cnode(2,1){0.4}{C}
\cnode(1,2){0.4}{D}

\cnode(5,3){0.4}{E}
\cnode(6,2){0.4}{F}
\cnode(5,1){0.4}{G}
\cnode(4,2){0.4}{H}

\cnode(8,3){0.4}{I}
\cnode(9,2){0.4}{J}
\cnode(8,1){0.4}{K}
\cnode(7,2){0.4}{L}
\cnode(11,3){0.4}{M}
\cnode(12,2){0.4}{N}
\cnode(11,1){0.4}{O}
\cnode(10,2){0.4}{P}

\ncline{B}{A}
\ncline{D}{A}
\ncline{C}{A}
\ncline{B}{D}
\ncline{B}{C}

\ncline{F}{G}
\ncline{F}{H}
\ncline{F}{E}
\ncline{E}{H}
\ncline{E}{G}
\ncline{H}{G}

\ncline{L}{I}
\ncline{L}{J}
\ncline{L}{K}

\ncline{I}{K}
\ncline{K}{J}

\ncline{N}{M}
\ncline{N}{O}
\ncline{N}{P}
\ncline{O}{P}
\ncline{P}{M}
\ncline{O}{M}

\end{pspicture}
\end{center}

\caption{A Bayesian network structure learned during the evolutionary process of an EDA for the concatenated trap-4 problem. An edge represents dependence among variables.}

\label{bayes_trap}
\end{figure}

On the other hand, an EDA based on Bayesian networks can capture the interactions among the genes:

\begin{equation}  {\pi}( \boldsymbol{x})=\prod_{i=1}^m {\pi}_i(x_i| {{\boldsymbol{pa}_i}} ) \label{bayes_net} \end{equation}

\begin{figure}[b]
\centering
\includegraphics[scale=0.6]{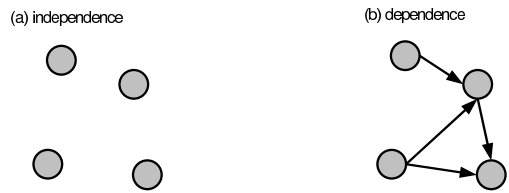}

\caption{(a) the model of independence among variables, as adopted by PBIL \cite{pbil} and cGA \cite{cga} (b) the model adopted by higher-order EDAs, which assumes that some dependence structure exists and should be inferred from the selected individuals.}

\label{estrut2}
\end{figure}

\begin{tabular}[ht]{rl}
where & $\pi$ is the joint probabilistic model\\
      &$\boldsymbol{x}$ is the binary vector of an individual\\
      & ${\pi}_i(x_i)$ is the marginal model of gene i\\
      & $\boldsymbol{pa}_i$ is the set of parents of $x_i$ \\
      &\\
\end{tabular}
\\
The set $\boldsymbol{pa}_i$ comprises the variables on which $x_i$ depends. This factorization is only possible because, for each $i$,  $x_i$ is assumed to be independent of its nondescendants, given its parents $\boldsymbol{pa}_i$ \cite{pearl88}.  Figure \ref{estrut2} illustrates two possible factorization assumptions and figure \ref{bayes_trap} illustrates a Bayesian network learned at some stage of an evolutionary process for the concatenated trap-4  problem \cite{pelik99} with 4 subproblems. 
EDAs based on higher order statistics (typically higher than order 2) should be called higher-order EDAs, in contrast to low-order EDAs which use statistical models where no interactions or, at most, only pairwise interactions are considered. Two very representative higher-order EDAs which adopt Bayesian Networks are the Bayesian Optimization Algorithm (BOA) \cite{pelik99} and the Estimation of Bayesian Network Algorithm (EBNA) \cite{ebna}. The difference among them is on the metric used to evaluate factorizations when searching for the model structure.


\section{Related work}

\label{related}

\begin{figure}[b]
\centering
\includegraphics[scale=0.6]{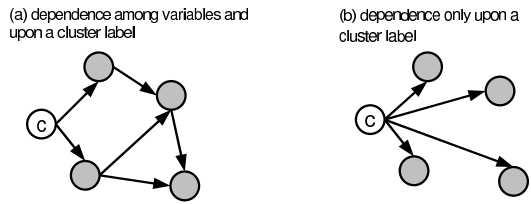}

\caption{(a) the model adopted by UEBNA, including a cluster label in the structure (b) the model adopted by a clustered order 1 EDA with the concept-guided combination. All variables are assumed to be independent, and the only dependence is upon the cluster label $C$. This model is adopted by the $\varphi$-PBIL algorithm}

\label{estrut1}
\end{figure}



It was already shown that clustering improves the performance of some EDAs for certain classes of problems. For globally multimodal problems, for instance, clustering has revealed a promising approach \cite{pena2005}. Such problems present several global optima or, in other words, several optimal solutions with the same fitness. Those problems may be very tricky for an evolutionary algorithm to solve, since slow convergence to one of the optima (genetic drift) often occurs. The retarded convergence is explained by the combination of solutions coming from different regions of the search space, which often results in poor solutions. The application of clustering in EDAs when solving globally multimodal problems can be described as the separation of the population in subpopulations (one for each cluster), and the subsequent learning of different probabilistic models for each subpopulation.  Breeding generally involves no combination among different subpopulations. Clustering improves the performance of the evolutionary algorithm by such an interbreeding avoidance. However, this approach leads to a smaller exploration of the space, since combination from distant regions could be favorable.

Clustering is adopted, for the first time, as the niching technique of an EDA in \cite{pelikan2000}. Only binary codifications are considered, and k-means clustering was applied to obtain subpopulations at each generation. Representative EDAs and widely known globally multimodal test problems were used in the experiments. The most noticeable result is the major improvement in effectiveness and convergence speed of UMDA -- a single-order EDA -- when clustering is applied. The clustered UMDA outperforms a simple UMDA for multimodal problems without a complex structure like twomax. The number of global optima found and stably maintained is increased with clustering. Those results are explained by the speciation of each cluster in a different peak. 

However, for more structured problems, clustering does not help UMDA very much. Actually, the literature does not report any low-order EDA which is able to solve hard deceptive structured problems, even when aided by clustering. Single-order clustered EDAs are only expected to perform well on globally multimodal optimization problems because combination of solutions from different basins is avoided \cite{pena2005}. Actually, the beneficial effect claimed by \cite{pelikan2000} is to control combination among different subpopulations
since interbreeding is considered as harmful for the evolutionary process. Exploration of the search space is done within each cluster by an evolutionary algorithm in a nearly parallel fashion. 

The authors of \cite{pelikan2000} conclude by stating that more sophisticated operators, besides just clustering, should be used for an efficient scale-up behavior on low-order EDAs on difficult problems. 




A similar approach to clustering is the parallelization of EDAs by adopting multiple subpopulations and migration. In \cite{P2BIL}, a simple recombination operator called PV-wise uniform crossover is proposed, which is similar to GA uniform crossover. After two parent PVs were selected for combination 
a new temporary PV is built by randomly selecting, for each gene, from which of the two parents should the binomial proportion be taken from. A new individual is sampled from this PV. 

It is possible to derive a general framework for a very simple EDA based on single order statistics and aided by some sort of clustering, as in \cite{pelikan2000}. Incrementally, or at each generation, the best individuals would be  clustered by similarity of their genotype and, for each cluster, independent models for each gene would be learned. Assuming a fixed number $k$ of clusters and a fixed number $p$ of genes, a total of $p\cdot k$ independent binomial proportions would be kept updated during the evolutionary process. New individuals would be generated by sampling from the probability vector of a single cluster, or combination among clusters (interbreeding), similarly as in \cite{P2BIL}. 


Another perspective for the application of clustering in EDAs can be found in \cite{pena2005}. The unsupervised estimation of Bayesian network algorithm (UEBNA) uses a model-based approach for clustering, based on the unsupervised learning of Bayesian networks at each generation. An unobserved variable $C$ is included in the model, which represents the unknown cluster label. The Bayesian network represents a joint probabilistic model for the best individuals, considering a factorization which includes all genes and the cluster random variable $C$. 


The structure of a Bayesian network describes a factorization as a directed acyclic graph, as illustrated in figure \ref{estrut2}{b}. The nodes denote random variables and the edges represent dependencies among those variables. Additionally, a set of parameters of the network specify conditional distributions given the possible values of the corresponding parents $\boldsymbol{pa}_i$ of each variable $i$. UEBNA adopts unsupervised learning of Bayesian networks since the cluster variable is unspecified. The structure of this kind of Bayesian network is illustrated in figure \ref{estrut1}{a}. 

The algorithm is expected to detect the correct setting of diverse global optima to diverse subpopulations and capture dependencies among the variables of the problem through learning Bayesian networks at successive generations. Experimental validation confirms the effectiveness of the algorithm for a set of globally multimodal structured problems. The most relevant problem tested was graph bisection, which is a hard problem even for higher-order EDAs like EBNA or BOA. Actually, UEBNA outperforms EBNA achieving a higher number of global optima especially for greater instances of graph bisection, which validates the relevance of clustering for evolutionary computation. 

Unfortunately, the Bayesian network-based model adopted by  EBNA, UEBNA, BOA and other higher-order EDAs requires a computationally complex learning stage, which must be run at each generation. Actually, finding the best structure of a Bayesian network is known to be a NP-hard optimization problem itself. In practice, a greedy algorithm is executed to perform the search, which can be based on penalized maximum likelihood. Penalization avoids excessively complex network structures. 

It was already recognized that learning the structure of a Bayesian network at each generation may become a bottleneck for EDAs and some attempts to overcome this problem have been proposed \cite{sporadic_model}. On the other hand, incorporating even more elements into the model, as in \cite{pena2005} where clustering labels are used, seems to be worthy and should be considered, but this approach results models of higher complexity.

It would be very worthy, however, if a simpler EDA could reach similar performance as more complex EDAs do. As suggested by \cite{pelikan2000}, clustering itself does not solve the problem, but further mechanisms could also be usefully incorporated. In the next sections a combination operator is evaluated, which allows a low-order clustered EDA to achieve success on the linkage learning task. This operator explores information gained from clustering itself and from single order statistics from the subpopulations. The proposed approach allows an effective exploration of the search space by combining the most relevant information. Experiments illustrate that the algorithm preserves the structure of complex problems which present high interactions among variables, despite only adopting statistics of low order. 

Two major motivations for the application of clustering in EDAs can be derived form the works reviewed above: (i) to increase preservation of the diversity in the population and (ii) to avoid combination from different peaks in multimodal optimization. Another motivation for clustering is (iii) to allow the identification of a mixture of distributions in continuous optimization. This last motivation is often described in the literature \cite{bosman2001}\cite{ahn2006}, but it is not directly related to the work herein presented.

The relevance of motivation (i) is clear since diversity maintenance prevents premature convergence to local optima and allows for a proper exploration of the search space. This topic is related to the niching methods of simple genetic algorithms, and was widely studied. Motivation (ii) is, however, conflictive with the first. The combination of information from distant regions is a strong mechanism for exploring the space, since it could potentially result in better solutions. Even combination from different local optima is useful, since those optima may result from different substructures, which were already found and should be combined. However, this aspect was not well studied yet and is explored later in this paper.

\section{Using an information measure to guide combination }
\label{interbreeding}

Low-order EDAs have not been effective on solving several classes of problems, particularlly when high order of interactions among the variables occur. Clustering was applied as a niching technique for EDAs, as reviewed in Section \ref{related}. Individuals are grouped by similarity of their genotype, and combination among different groups is usually avoided. However, the performance of simple low-order EDAs was not much increased by using clustering. 




One of the reasons which motivate the application of niching is to avoid premature convergence to local optima. Diversity preservation, however, is also important even when a single optimum exists \cite{mahfoud} since substructures should be explored and recombined before the global optima is correctly identified. 

The existence of substructures in the problem is also an important factor which motivates niching. When an evolutionary algorithm is relatively away from convergence, we can conjecture that the main source of diversity in the population is due to the diversity of substructures, since global optima are not clear yet. Thus, at some stage of the process, a clustering algorithm would group together individuals possessing the same substructures or, in other words, different substructures would characterize each subpopulation. 

The concepts which characterize each cluster are, thus, relevant information which should be explored. Actually, the combination of important parts of promising solutions found so far should guide GAs and EDAs during the exploration of the search space \cite{pelikan2000}. Therefore, combination from different clusters should not be avoided, but carefully performed. 

This Section describes a combination mechanism for clustered EDAs which carefully chooses how to combine information from clusters. Section \ref{related} reviewed a random naive combination mechanism called PV-wise crossover \cite{P2BIL}. However, it is shown later that this operator is not effective for the exploration of the search space and is not able to combine valuable information from parents, similarly as the GA with uniform crossover. 

A more sophisticated but still very computationally efficient operator can be built using ideas from information theory. It also combines information from different clusters by building a temporary PV from two randomly selected parent PVs. The difference is on how to choose from which parent to take each binomial proportion. A measure from information theory guides the choice of the best parent for each gene. It aims for a careful combination of relevant information from two PVs, attempting to maintain the most informative part of each parent. This operator is called concept-guided combination or, for brevity, the cg-combination. 


\begin{figure}[t]
\centering
\includegraphics[scale=0.4]{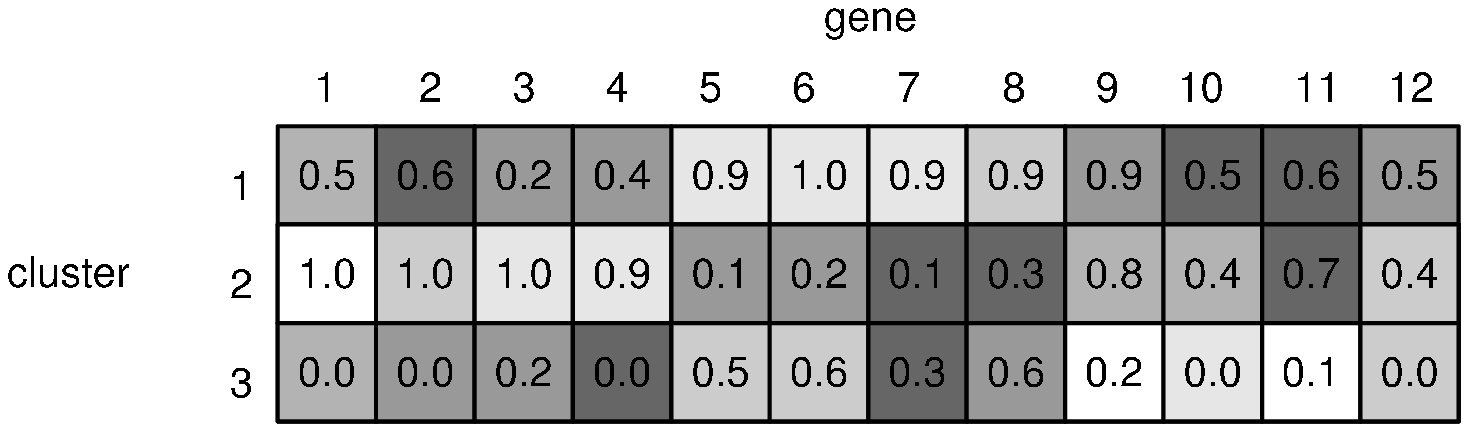}

\caption{A possible setting of binomial proportions $\hat{\pi}_{i,j}$s and their respective $\hat{w}_{i,j}$s at some stage of the evolutionary process for a certain artificial problem. The values inscribed are the $\hat{\pi}_{i,j}$s, while the grey tones denote the $\hat{w}_{i,j}$ in a scale ranging from black (minimum) to white (maximum).}

\label{binomiais}
\end{figure}

\begin{figure}[b]
\centering
 \includegraphics[scale=0.5]{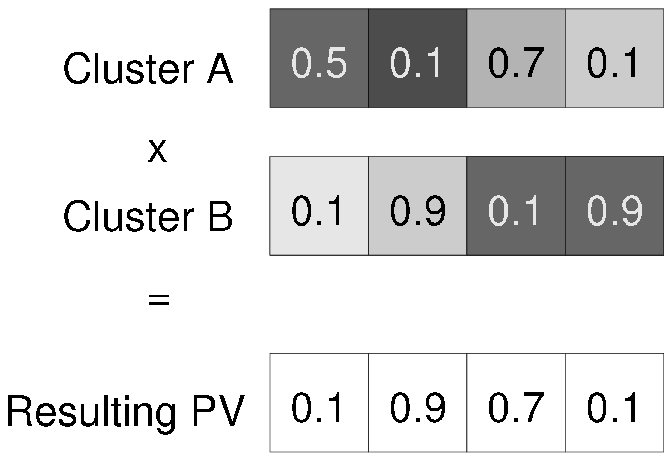}

\caption{Generating a temporary PV during an interbreeding operation. $\hat{\pi}_{A,.}$, $\hat{\pi}_{B,.}$ and their respective $\hat{w}_{i,j}$s are shown using the same color convention as in figure \ref{binomiais} (except for the resulting PV which has no $\hat{w}_{i,j}$s). }

\label{pvs}
\end{figure}

During a cg-combination, two parent clusters -- $A$ and $B$ --  are, randomly and proportionally to the mean fitness, selected. A temporary PV, from which a single new individual will be created, is obtained by taking proportions for each position $j$ from either $\hat{\pi}_{A,j}$ or $\hat{\pi}_{B,j}$. A measure from information theory is used to guide that choice in order to select always the most informative parent for each gene. Let      
\begin{equation} \hat{h}_{\cdot,j}= -\sum_{q\in \{0,1\}}{\hat{p}_{j,q} log_{2}(\hat{p}_{j,q})}\end{equation} be the entropy of the distribution of gene $j$, where  $\hat{p}_{j,q}=P(x_j=q)$ is the proportion of individuals possessing the value $q$ for gene $j$ in the whole population ($q\in \{0,1\}$). Similarly, let \begin{equation} \hat{h}'_{i,j}= -\sum_{q\in \{0,1\}}{\hat{p}'_{i,j,q} log_{2}(\hat{p}'_{i,j,q})}\end{equation} be the entropy of the distribution of the same gene $j$ without taking cluster $i$ into account in the estimated proportions.   The proportion $\hat{p}'_{i,j,q}$ is computed as $P(x_j=q\mid c(\boldsymbol{x})\in \{1,2,\ldots,k\}-\{i\})$.

The measure of how informative is a cluster $i$ to a gene $j$ is therefore the difference in the entropy of the distribution of the gene $j$ before and after observing cluster $i$:

     \begin{equation}\hat{w}_{i,j}=\hat{h}_{i,j}-\hat{h}'_{i,j}\end{equation}

Thus the decision of how to build the temporary PV becomes simple: choose, for each gene $j$, the parent with the greatest $\hat{w}_{i,j}$ among all $i$s. Each position $v_j$ of the temporary PV $v$ is defined as 

\begin{equation} v_j=\begin{cases} 
\hat{\pi}_{A,j},\:\:if \: \hat{w}_{A,j} > \hat{w}_{B,j} \\
 \hat{\pi}_{B,j},\:\:otherwise \end{cases} \end{equation}

After computation of $v$ a new individual is generated by sampling from each position of $v$ independently. Figure \ref{binomiais} shows a scenario with $k=3$ clusters and 12 variables. Figure \ref{pvs} illustrates the creation of $v$ after an interbreeding for a small problem with four variables. 

The cg-combination relies that clustering could group together individuals possessing the same substructure, and attepts to extract the information about that substructure from $\hat{W}$ and $\hat{\Pi}$. A given cluster is expected to be informative for all genes which define a building block if most individuals in that cluster possess that building block.  Overlapping building blocks do not represent a problem: two clusters would be informative to the same gene and this would still preserve substructures. This behavior is illustrated in Section \ref{experimentos}. 

In addition, high-fitness similar individuals resulting form combinations of lower-level building blocks would also be grouped together. Thus, relevant substructures are combined as the process goes on. 

The next Section presents a low-order EDA that effectively adopts the cg-combination as the interbreeding mechanism.

\section{$\varphi$-PBIL: A Clustering-based Evolutionary Algorithm}





The previous Section presented an operator called cg-combination, which supports the combination of information from different clusters and, potentially, allows a low-order clustered EDA to effectively perform linkage learning. 
This operator can be used in the ``interbreed'' step. 
Each cluster defines a probabilistic model and genes are assumed conditionally independent given the cluster label, therefore only dependencies between a gene and the cluster label are considered. The resulting model is, therefore, limited to order 2 statistics.

The algorithm $\varphi$-PBIL (Concept-guided Population-Based Incremental Learning) \cite{EP07a} is a low-order EDA which follows the proposed approach and performs interbreeding. Algorithm \ref{algo} shows the pseudocode of $\varphi$-PBIL. 

\begin{algorithm}[h]
\caption{$\varphi$-PBIL}
\label{algo}

\begin{algorithmic}

\STATE \textit{// Initialization}: \\
\STATE Generate an initial random population of size $N_0$. \\
\STATE Compute the fitness of the individuals. \\
\STATE Select only the $N_w (N_w<N_0)$ best. \\
\STATE \textit{// Learning}: 
\STATE Learn clusters from the population.\\ 
\STATE Compute a probability vector (PV) of binomial proportions for each cluster.\\
\STATE Store the PV's in the matrix $\hat{\Pi}=(\hat{\pi}_{i,j})$.\\
\STATE Compute the matrix of information measures $\hat{W}=(\hat{w}_{i,j})$.\\
\WHILE{convergence criteria are not met}
\STATE  \textit{//  Breed a new individual $H$:}\\
\STATE Create $H$ by randomly choosing from one of the two following possible procedures:\\
\STATE \hspace{5pt}(i) sample from one of the PVs \\
\STATE \hspace{5pt} or (ii) apply interbreeding, guided by the matrix\\ 
 \hspace{17pt}$\hat{W}$, and sample from the combined PV. \\
\STATE \textit{//  Selection}: 
\STATE Compute the fitness of the new individual $H$. 
\IF {$H$ is not worse than the worst in population}
\STATE Delete this worst individual. 
\STATE Insert $H$ in the population.
\ENDIF
\STATE \textit{// Learning:}
\STATE Update clusters and matrixes $\hat{\Pi}$ and $\hat{W}$.\\
\ENDWHILE

\end{algorithmic}
\end{algorithm}


$\varphi$-PBIL follows an incremental architecture where a single individual is generated at each iteration without the adoption of a succession of ``generations'' as other EDAs often do. A fixed number of $k$ clusters are maintained and continuously updated.  Whenever a new individual is generated, it replaces the worst individual in the population (if the new individual is not even worse) and, subsequently, clustering hypothesis and probabilistic models for each cluster are just updated through a single typical k-means step, and not fully relearned. Each cluster defines a subpopulation and, since only binary variables are allowed, then the probabilistic models for the subpopulations are just the corresponding binomial proportions $\hat{\Pi}=(\hat{\pi}_{i,j})$ which denote the proportions of individuals with the value 1 for each gene $j$ on each cluster $i$. 

Sampling from one of the probability vectors (PVs) will generate a new individual, similarly to other clustered EDAs. A PV is chosen randomly and proportionally to the mean fitness of the individuals of each corresponding cluster. 

Sampling from a single PV, however, is not the only option for generating a new individual in $\varphi$-PBIL, since the combination of PVs is also considered. Actually, the cg-combination is chosen as the default interbreeding mechanism for $\varphi$-PBIL. Two interbreeding mechanisms were discussed in the previous Section. Both propose to combine two PVs, obtain a temporary PV $v$ and sample from $v$ to generate a new individual. Experiments in the next sections illustrate how the PV uniform crossover fails on combining relevant information from two clusters, while the cg-combination succeeds.

Two features were added in order to increase the performance of the algorithm. The first feature was motivated by the need for local search. The maximum likelihood estimator (MLE) of a binomial proportion is the simple mean of successes. In our case, it should be interpreted as the proportion of ones at each gene, or \begin{equation} \hat{\pi}_{i,j}=\frac{\sum_{c(x)=i}{x_j}}{n_i} \label{eqmle}\end{equation} where $\sum_{c(x)=i}{x_j}$ is the number of individuals in cluster $i$ possessing the value $1$ at gene $j$ and $n_i$ is the number of individuals in cluster $i$. Unfortunately, when all individuals in a cluster have the same value (all 0's, or all 1's) in a certain locus, this estimator saturates at one of the extremes ($0\%$ or $100\%$). Sampling from those extreme proportions wipes out the chance of the alternative value, 1 or 0, to be generated and this behavior is harmful for local search.

\begin{table}[t]
\begin{center}
\begin{small}
     \caption{$\varphi$-PBIL parameters}

\begin{tabular}[ht]{|c|c|c|} 
\hline
Parameter&Description&Default \\
&&value\\
\hline
$N_0$ &initial population size & -- \\ \hline
$N_w$ & working population size & -- \\ \hline
$k$ & number of clusters  & -- \\ \hline
$p_c$  & probability of interbreeding&50\%\\ \hline
$p_{old}$ & probability of an old & 50\%\\ 
& clustering hyposthesis to be used & \\ \hline
$p_{w}$ & probability of the Wilson& 50\%\\ 
& estimator to be used& \\ 

\hline 
\end{tabular}

\label{param}
\end{small}
\end{center}
\end{table}

A perturbation mechanism was added which changes slightly the binomial proportions estimated and, therefore, allows for an allele to be generated even if all individual possess the complementary allele. The Wilson estimator reviewed in  \cite{agresti} incorporates a degree of uncertainty by estimating the binomial proportion as 

\begin{equation} \hat{\pi}_{i,j}=\frac{\sum_{c(x)=i}{x_j}+2}{n_i+4}\end{equation} is used instead of (\ref{eqmle}), which is the maximum likelihood estimator (sample mean).

Thus, for instance, even if all 100 individuals in a cluster possess the value 1 in a locus there still remains a probability of 2\% of a new individual to be generated with a 0 in that same locus. Some kind of mutation operator could also be applied, probably with similar effects on local search. Bayesian inference could also be applied since it properly express the uncertainty about the value of the parameter through the specification of a prior distribution for the parameter. 

Another feature was added in order to improve recombination for overlapping and hierarchical building blocks. When smaller BBs start blending, new BBs emerge. This causes the loss of older clustering hypotheses, consequently inhibiting those initial BBs in recent individuals. This is a consequence of the well known race between selection and innovation \cite{gold98}. For overlapping and/or hierarchical building blocks, this constitutes a great problem. A potential solution is the maintenance of some old BBs (therefore, old clustering hypotheses), which can be used to generate new individuals. Both the current and an old clustering hypothesis compete when breeding. This mechanism works as follows: both $\hat{W}=(\hat{w}_{i,j})$ and $\hat{\Pi}=(\hat{\pi}_{i,j})$ matrixes are stored, for some old well-performing clustering hypotheses. The cg-combination selects randomly from one of two clustering hypotheses: the old set $\{\hat{\Pi}_{old},\hat{W}_{old}\}$ and the new, recently updated set $\{\hat{\Pi},\hat{W}\}$. It also computes performance records about both sets: if the new individual, generated by one of the sets, is selected, then the performance variable of the corresponding set is updated (increased by one). If the new set overperforms the old one, then the old set and its performance variable are updated, and the performance variable of the new set is set to zero. 


   \begin{figure*}[]
\begin{center}
     {\includegraphics[scale=0.8]{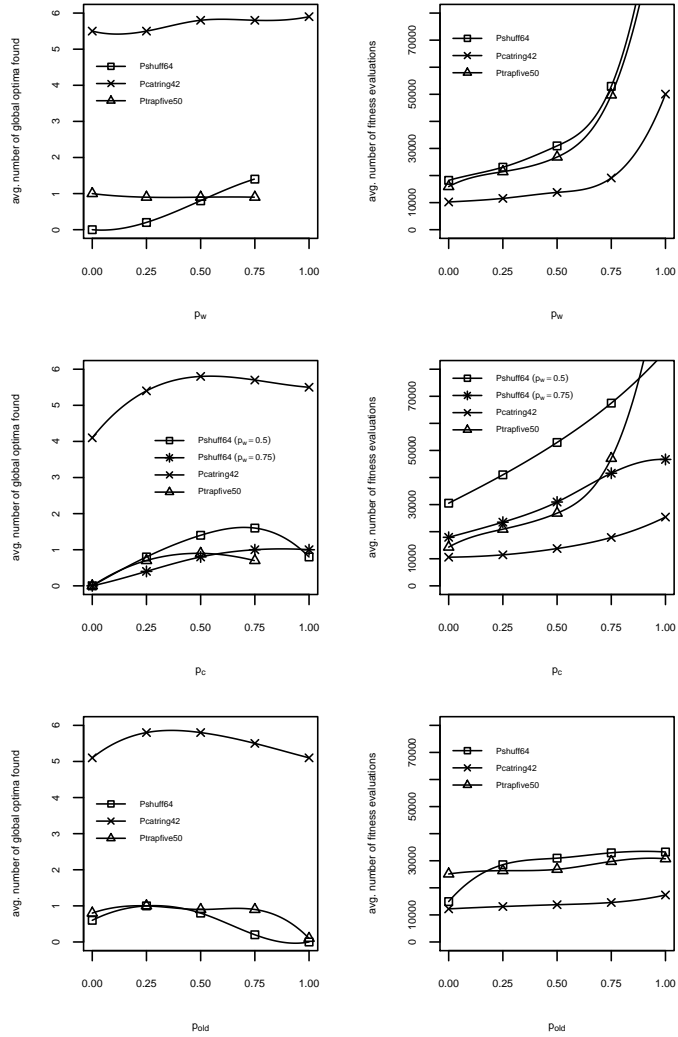}}
    \caption{ Empirical evaluation of the sensitivity of $\varphi$-PBIL to three parameters $p_c$, $p_{old}$ and $p_{w}$, reporting the mean number of global optima found and maintained and the mean number of fitness evaluations until convergence. Five values for each parameter (0\%, 25\%, 50\%, 75\% and 100\%) are considered on each row. Assume $p_c=50\%$, $p_w=50\%$ and  $p_{old}=50\%$ where the value of the parameter is not explicit. Each point in the graph represents the mean of 10 independent runs of the algorithm (replications). Curves result from spline interpolation. When the mean number of fitness evaluations exceeds 100,000 the respective line segment in the graph for the number of global optima is truncated.
}

    \label{fig_param}
\end{center}
    \end{figure*}

All parameters for $\varphi$-PBIL are described in table \ref{param}. Some default values for most parameters were set after empirical investigation, which is shown in Section \ref{sect_param}. $p_c$ (probability of interbreeding), 
$p_{old}$ (probability of an old clustering hypothesis to be used during breeding instead of the current one) and $p_{w}$ (probability of the Wilson estimator to be used instead of the sample mean) are all set to $50\%$ by default. 

Termination criterion of the algorithm was set to be the loss of diversity inside PVs: the algorithm finishes when all $\hat{\pi}_{i,j}$s saturate (reach some value above $0.95$ or below $0.05$). This is said to be the condition for the convergence.

The user must set the values for some parameters for which no default values are provided: initial population size $N_0$, working population size $N_w$ and the number of clusters $k$. The working population exists after the initialization and is created from the best individuals of the initial population as described in algorithm \ref{algo}.

The next Section describes an empirical investigation performed to find a default parameter setting for the algorithm.

\section{Parameter setting}

\label{sect_param}

This Section shows an exploration in the parameter space using three representative test problems: shuffled HIFF \cite{hiff}, concatenated trap-5 \cite{pelik99} and graph bisection \cite{pena2005}. The experiment aims to find default values for some parameters of $\varphi$-PBIL. We start from a hypothesis that $p_c$, $p_{old}$ and $p_{w}$ should be all set to $50\%$. The problem instances, which are descrbed in the next section) are Pshuff64, Ptrapfive50 and Pcatring42, all three relatively small (64, 50 and 42 variables, respectively), but are shown to be enough for revealing an influence of the parameters on the performance of the algorithm. Pcatring42 and Pshuff64 are  globally multimodal, possessing 6 and 2 global optima respectively, while Ptrapfive50 has a single global optimum.

The other three parameters are maintained fixed for all runs of each problem in this experiment: $N_0=2,500$, $N_w=250$ and $k=10$ for Pcatring42 and $N_0=3,000$, $N_w=300$ and $k=15$ for Pshuff64 and Ptrapfive50. Those values were previously verified to be near the minimal values at which the algorithm becomes able to find all global solutions for each of the three instances considered here in this evaluation.

Figure \ref{fig_param} shows the results of this investigation. Each graph shows the variation of one of the three parameters considered, where the other two stay fixed. When nothing else is stated, the value for the remaining two considered parameters is $50\%$. 

Changing one of the parameters can dramatically affect the behavior of the algorithm for one or more of the problems. Setting $p_{old}$ to $0\%$ or $100\%$ causes a reduction on the performance on Pshuff64 since both old and new clustering hypothesis store information about the subsequent levels of the problem structure. The performance on other problems is, however, not so influenced by $p_{old}$, except when $p_{old}$ is at $100\%$, where performance on Ptrapfive50 also drops down. Using only old clustering hypothesis seems to be harmful for the algorithm.

Setting the probability of interbreeding -- $p_c$ -- to zero affects the performance of the algorithm for a wider range of problems, since this mechanism is responsible for recombination. 
Increasing $p_c$ slows down the convergence for all instances tested.

The parameter $p_w$ is also negatively related to the convergence speed, but lower values for $p_w$ should be avoided, at least for HIFF, since no global solution for Pshuff64 is found for $p_w=0$. The effectiveness on other problems was not affected by this parameter.

After analyzing all results and recognizing that nearly extreme values for all of the parameters tested are, often, undesirable, therefore we set $p_c$, $p_{old}$ and $p_{w}$ all to $50\%$.

\section{Some benchmark optimization problems}

This section discusses and revises the benchmark problems used in the empirical evaluation Section. The reader is referred to \cite{pena2005}\cite{hiff}\cite{pelik99}\cite{yu2005} for a more detailed description of the problems. 

A representative set of benchmark problems was chosen, which are known to be hard for a GA and most EDAs to solve. The existence of building blocks, or a structure in the problem, generally prevents GA and low-order EDAs from achieving success, since genes are assumed independent and structure is not preserved. Simple GA with one-point crossover and a proper encoding performs well for some structured problems, but the best encoding requires previous knowledge about the problem structure, something that is not available in general.

Three general classes of structured problems are considered. The simpler class is that of additively decomposable functions (ADFs).  In ADFs there is no dependence structure among the subproblems since the contribution of each substructure to the overall fitness of the solution is independent of the value of the remaining variables, therefore all subproblems can be solved independently. The concatenated k-trap function \cite {pelik99}, which is better described below, is an example of an ADF with $l/k$ subproblems, where $l$ is the size of the problem and $k$ is the size of each subproblem. The highest order of dependence in the concatenated k-trap is, therefore, $k$. Most ADFs proposed in the literature, including trap-5, are deceptive. This means that lower order statistics would lead the algorithm away from finding the correct building block, and motivates the use of higher order statistics. 

A slight variation of ADFs is obtained when overlapping building blocks exist but the problem structure is still defined by additive functions.  When building blocks overlap, some genes are present in more than one subproblem simultaneously and the problem structure learning task is harder. This class of problems should be called overlapping additively decomposable functions  (OADFs). \cite{yu2005} discusses how to design combination strategies for GA when overlapping building blocks occur, in order to minimize the number of building blocks disrupted. 

Another class of GA-hard problems comprises Hierarchically decomposable functions (HDF). They were proposed after a criticism to ADFs, which present no dependence structure among the subproblems. HDFs, in turn, are much harder to optimize since the substructures interact. The fitness contribution of two substructures, when combined, is different from the sum of their individual contributions. One of the most important HDFs is HIFF \cite{hiff} which presents interactions up to the size of the problem. HDFs illustrate a more general class of problems than ADFs and demand a more complex mechanism for combination. Substructures of each level must generally be first identified and recombined before the next level starts emerging. 

Apart from the taxonomy above, two other important aspects of the benchmark problems recently approached in the literature should be mentioned. Problems that present several global optima, called globally multimodal problems, represent a source of difficulties for GAs and EDAs and require efficient niching mechanisms. Also, a class of symmetrical multimodal problems has recently been attracting attention. Symmetry occurs when some regularities on the landscape lead to the existence of complementary solutions with identical fitness and, consequently, to complementary global solutions. Globally multimodal and/or symmetrical problems can be considered hard for most evolutionary algorithms because combining solutions from two symmetric local optima will probably not be useful to find the global optima. 

Next, the problems used in this work are briefly stated.
\\




\subsubsection*{Twomax} 

Twomax is a simple function of $n$ binary variables:

	\begin{equation} f_{twomax}(\textbf{Z})= \left|\frac{n}{2}-\sum_{i=1}^{n}{z_i} \right| \end{equation}

There are two global maxima: $(0,0,\ldots,0)$ and $(1,1,\ldots,1)$, both with fitness equal to $n/2$. \\

\subsubsection*{Concatenated trap-5}

The concatenated trap-5 \cite{pelik99} is an additively decomposable function which possess a single global optima (the string fulfilled with 1s), with a fitness equal to the size of the problem. The input string is partitioned into disjoint groups of five bits each. A 5-bit trap function is applied to each of the groups and the fitness of the individual is the sum of the contributions of each 5-bit group. The algorithm should be able to learn and preserve that structure in order to achieve success. The 5-bit trap function is:

\begin{equation} trap_5(u) = \begin{cases} 5, \:\:\: if\:\: u=5\:\:, \\
                            4-u \:\: otherwise
              \end{cases} \end{equation} where u is sum of the bits in the group.

For each contiguous non-overlapping block of five variables, two building blocks can be identified: $(0,0,0,0,0)$ and $(1,1,1,1,1)$, which contribute with $4$ and $5$ to the overall fitness, respectively. Single order (or any lower than 5-order) statistics may lead apart from the optimal value for each gene, what makes this problem deceptive.

Notice that several combinations of optimal and suboptimal building blocks may lead to multiple suboptimal solutions. Actually, since there are 2 building blocks for a partition size of 5, then for a problem instance with size p, there are $2^{p/5}-1$ local optima and only one global optima, whose fitness is very close to that of the second best optima. An instance of size 50 (Ptrapfive50) is considered later.\\

\subsubsection*{Overlapping concatenated trap-5}

\begin{figure}[b]
\centering
\includegraphics[scale=0.8]{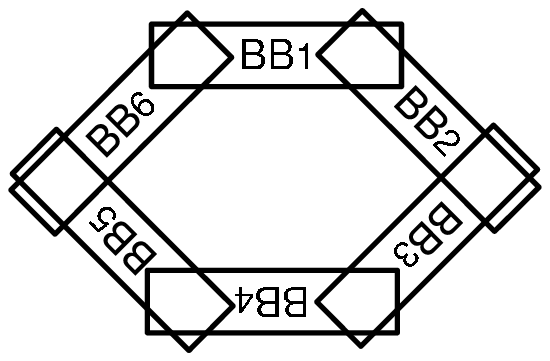}
\caption{A small instance of the overlapping concatenated trap-5 problem, with 6 building blocks. Overlapping length is 2.}

\label{figover}
\end{figure}

This problem is an overlapping additive decomposable function, with a fixed overlapping length. The overlapping scheme is circular as figure \ref{figover} illustrates for a problem with 6 BBs. Our test instance has 60 genes and overlapping length 2 (therefore 20 overlapping building blocks) called Poverfive60, which is also considered in \cite{yu2005}. The circular nature of the overlapping scheme with length 2 means that every building block shares 4 genes (2 with each neighbor) and the first building block is the neighbor of the last. In figure \ref{figover}, for instance, building block 1 shares 2 genes with each neighbor, building blocks 6 and 2. 

This problem is considered harder than the previous, concatenated trap-5, because in the overlapping version subproblems are not separable. \\

\subsubsection*{Hierarchical if-and-only-if} 
HIFF \cite{hiff} is an hierarchically decomposable function. A binary string of size $2^p$ represents a solution, where $p$ is the number of levels in the hierarchy. The fitness of a solution is given by 

\begin{equation} f_{hiff}(\textbf{\textit{B}}) = \begin{cases} 1,  \: if \: \left|\textbf{\textit{B}}\right|=1 \\ 
                              \left| \textbf{\textit{B}} \right| + f(\textbf{\textit{B}}_L) + f (\textbf{\textit{B}}_R), 
                               \\ \: \: \: \: if \: \left|\textbf{\textit{B}}\right|>1 \:and\: (\forall i \{ b_i = 0\}\:or\:\forall i \{b_i = 1\}) \\
                               f(\textbf{\textit{B}}_L) + f (\textbf{\textit{B}}_R),\:otherwise
                 \end{cases}\end{equation} where $\textbf{\textit{B}}$ is a block of bits  $(b_1,\ldots,b_{\left|\textbf{\textit{B}}\right|})$, $\left|\textbf{\textit{B}}\right|$ is the size of the block, $b_i$ is the i-th element in block. $\textbf{\textit{B}}_L$ and $\textbf{\textit{B}}_R$ are the left and right halves of the block $\textbf{\textit{B}}$ . The evaluation starts with the chromosome as a block.

Since tight linkage occurs, HIFF is relatively simple for a GA with one-point crossover to solve. In the shuffled version, however, the variables are randomly rearranged in order to avoid tight linkage. This version is much harder for a GA and some mechanism for detecting the problem structure is recommended. An experiment with the shuffled version of HIFF with 64 variables (Pshuff64) is described later. For EDAs in general the ordering of the genes in the codification is not relevant, since the probabilistic models do not take any advantadge from this information, as opposed to GA which actually may depend on an adequate codification.
\\

\hspace{1cm}

\subsubsection*{Graph Bisection} 

\begin{figure*}[h]
\centering
\includegraphics[scale=0.4]{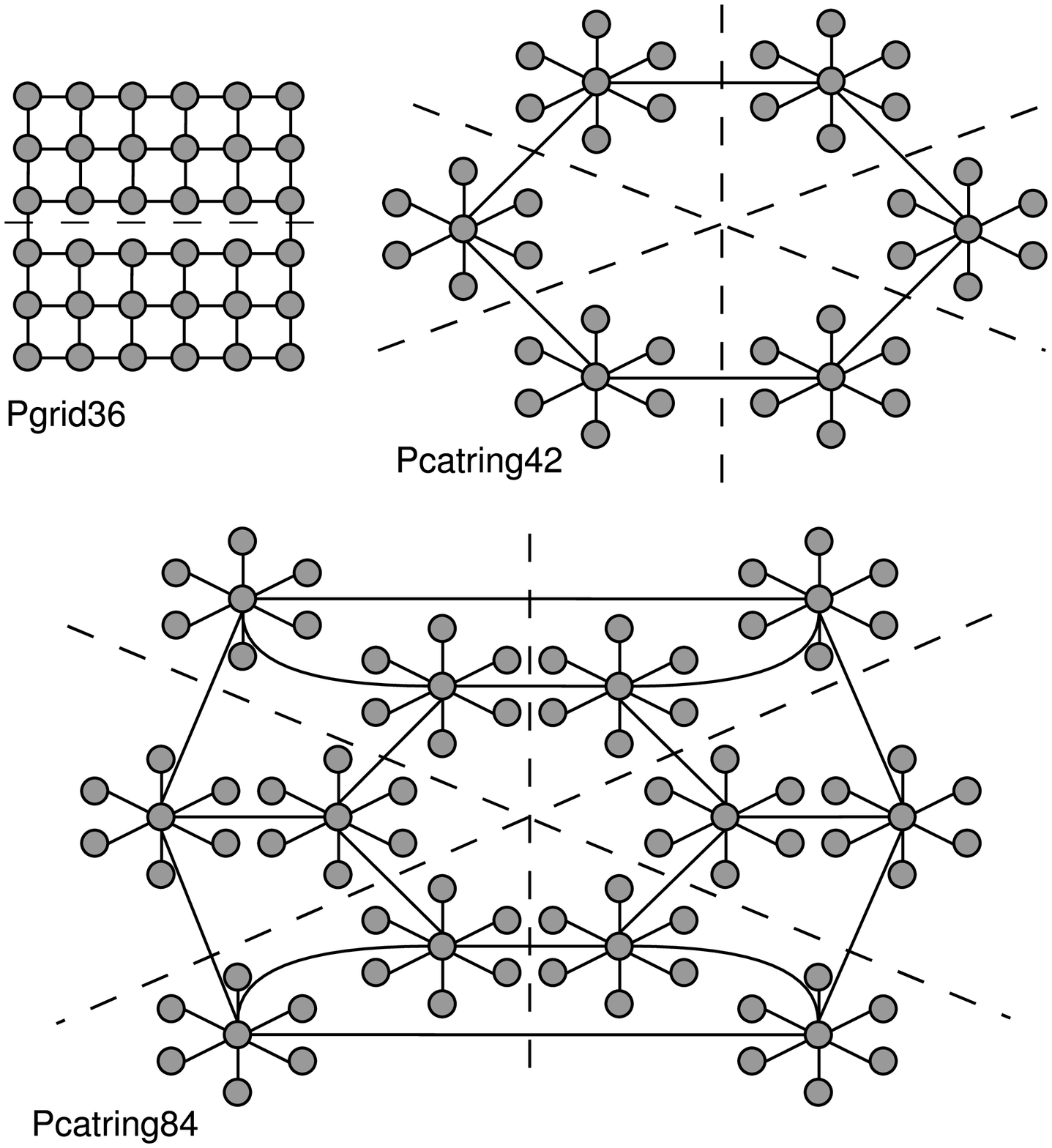}
\caption{ Graph topologies for Pgrid36, Pcatring42 and Pcatring84. Optimal partitions are illustrated by dashed lines.}
\label{topologia}
\end{figure*}

\begin{figure*}[h!]{\includegraphics[scale=0.91]{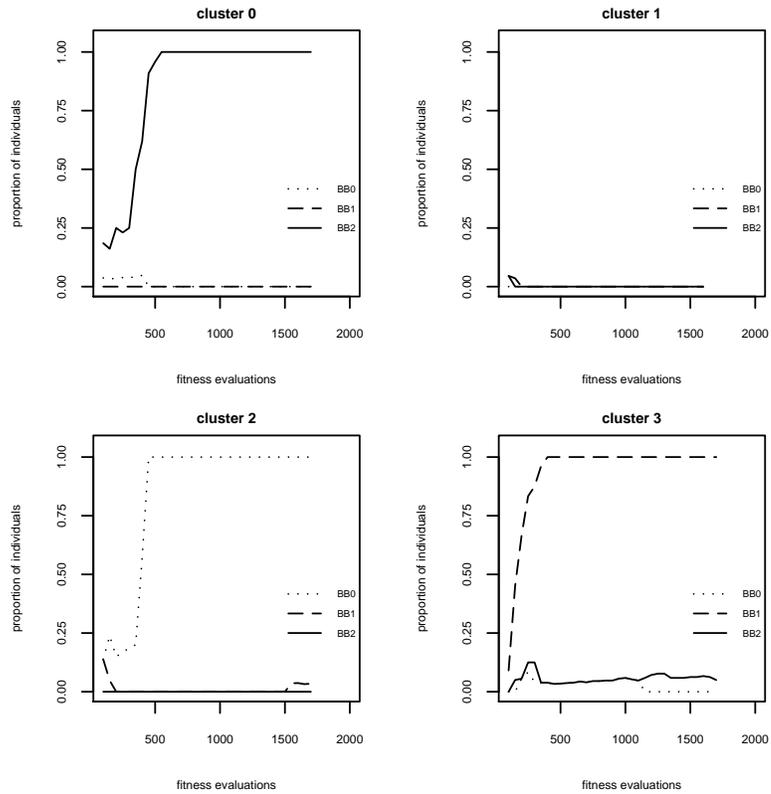}}
    \caption{A single run of $\varphi$-PBIL with the PV uniform crossover operator for a problem with 3 separable building blocks (BB0, BB1 and BB2). The number of fitness evaluations is reported in the figure since the process is incremental and there are no explicit ``generations''. Using PV uniform crossover, the algorithm fails on finding the global optima -- the combination of BB0, BB1 and BB2.\hspace{1cm}}
    \label{bad_clusters}

    \end{figure*}

\begin{figure*}[h!]

     {\includegraphics[scale=0.91]{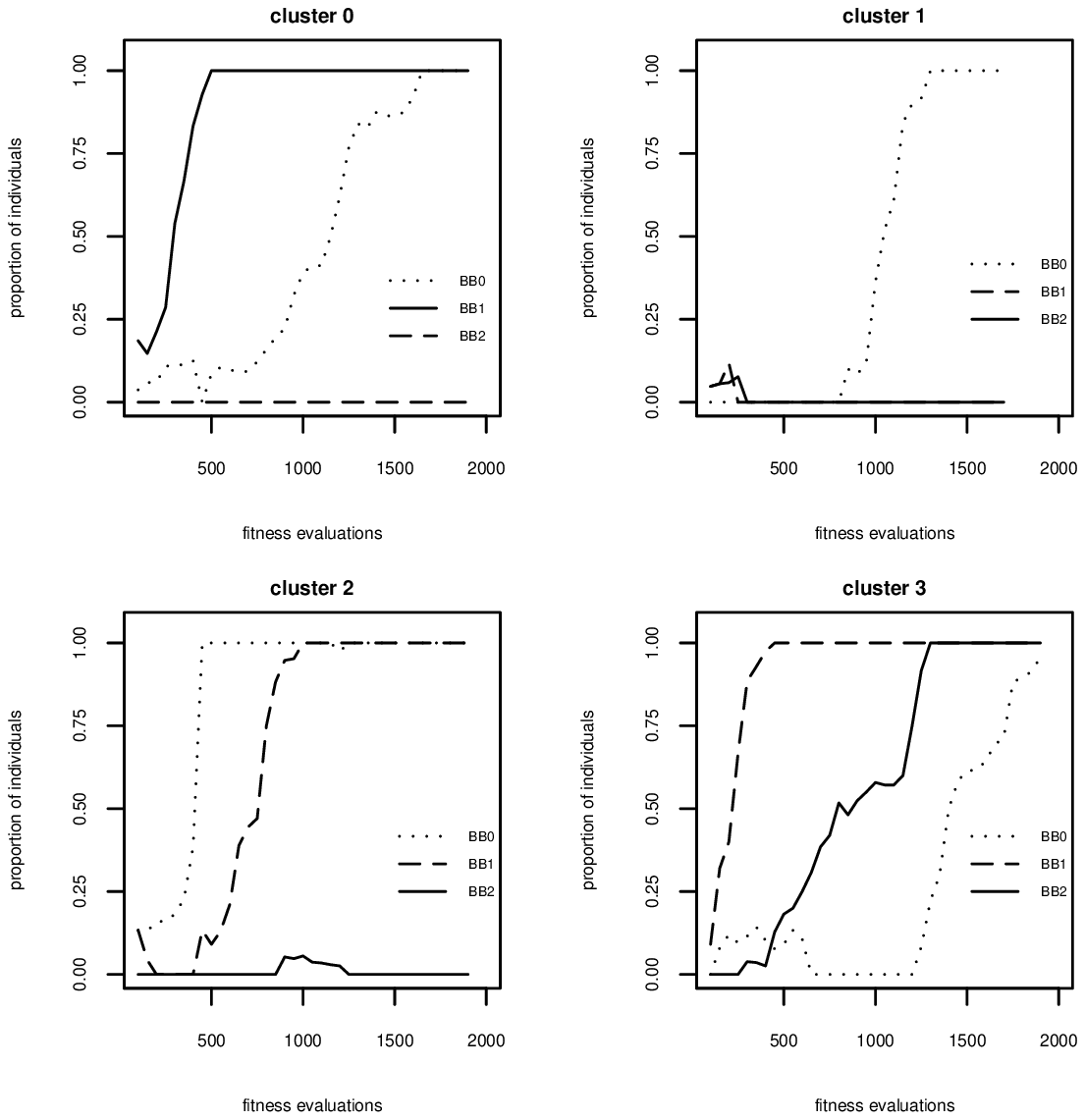}}
    \caption{A single run of $\varphi$-PBIL with the cg-combination for a problem with 3 separable building blocks (BB0, BB1 and BB2). The number of fitness evaluations is reported in the figure since the process is incremental and there are no explicit ``generations''. The algorithm successfully finds the combination of BB0, BB1 and BB2.\hspace{1.5cm}}
    \label{good_clusters}

    \end{figure*}

The objective of the graph bisection problem is to obtain two equally sized sets of nodes from the original graph such that the number of vertices linking both sets is minimal. The fitness of a given solution is the number of nodes minus the number of links between both sets, making it a maximization problem. The codification adopted is based on a binary vector of size n where the i-th gene represents the label of the i-th node. 

The set of possible solutions is restricted to the condition of equally sized sets; therefore, some individuals may appear which are unfeasible. A repair operator was adopted here and in \cite{pena2005} which randomly inverts the value of some genes. Iteratively a randomly selected gene in the majority is inverted until a feasible solution is obtained.

All instances of the graph bisection considered in \cite{pena2005} are also studied here. Figure \ref{topologia} illustrates three of those instances: Pgrid36 and Pcatring42 and Pcatring84, which possess 2, 6 and 6 global optima, respectively. The other instances considered are Pgrid16, Pgrid64, Pcat28, Pcat42, Pcat56, Pcatring28, Pcatring56 and Pcatring84, with the number of global optima ranging from 2 to 6.

\section{Empirical Evaluation}
\label{experimentos}

\begin{table}[h]
\label{tab_bommau}
\begin{center}
\begin{small}
\caption{Comparing the concept-guided combination with PV uniform crossover. The same parameters and random seeds were used in both sets of runs. A total of 30 runs were performed for each algorithm on each problem instance. The fraction of successful runs (Success \%), where at least one global optima is found, and the mean number of fitness evaluation (Eval.) until convergence and respective standard deviation (sd.) are reported. 
}

\begin{tabular}[b]{|@{\hspace{2pt}}l@{\hspace{1pt}}|r|r@{\hspace{4pt}}|} 
\hline
 &\multicolumn{1}{l|}{$\varphi$-PBIL using}&\multicolumn{1}{l|}{$\varphi$-PBIL using PV}\\
problem (size) & \multicolumn{1}{l|}{concept-guided}&\multicolumn{1}{l|}{uniform }\\
& \multicolumn{1}{l|}{combination}& \multicolumn{1}{l|}{crossover}\\
\cline{2-3}
& Success \% &Success \% \\
\hline
Shuffled HIFF (128)& 97\%	& 0\% \\
Concatenated trap5 (100)& 100\%& 0\%	\\
Overlapping c. trap5 (60)&100\%&0\%	\\
Twomax (100)&100\% &100\%\\
\hline

 \end{tabular}

\begin{tabular}[b]{|@{\hspace{2pt}}l@{\hspace{1pt}}|r@{$\pm$}@{\hspace{3pt}}l|r@{$\pm$}@{\hspace{3pt}}l@{\hspace{4pt}}|} 
\hline
 &\multicolumn{2}{l|}{$\varphi$-PBIL using}&\multicolumn{2}{l|}{$\varphi$-PBIL using PV}\\
problem (size) & \multicolumn{2}{l|}{concept-guided}&\multicolumn{2}{l|}{uniform }\\
& \multicolumn{2}{l|}{combination}& \multicolumn{2}{l|}{crossover}\\
\cline{2-5}
 &Eval.&sd& Eval.&sd \\
\hline
Shuffled HIFF (128)& 105134& 11612	& 81646 & 13474 \\
Concatenated trap5 (100)& 90474	&7203	&49391	&7366\\
Overlapping c. trap5 (60)& 55649	&4207	&23729	&1048\\
Twomax (100)&4825	&216	&4867	&318\\
\hline

 \end{tabular}
\end{small}
\end{center}

\end{table}

This Section illustrates the performance of the cg-combination and the $\varphi$-PBIL algorithm for a representative set of benchmark problems. Those problems summarize most of the hurdles that have been reported for benchmark problems for EDAs: deception, multimodallity, overlapping building blocks, global multimodallity and symmetry.

Initially, the two interbreeding mechanisms which were described in section \ref{interbreeding} are compared. 
Next, a comparison is performed between $\varphi$-PBIL and UEBNA \cite{pena2005}, which is a state-of-the-art EDA, both solving several instances of 
the graph bisection problem. \\

\subsection*{Comparing two interbreeding mechanisms}


These first experiments illustrates the behavior of the cg-combination, which is compared to the PV-wise uniform crossover \cite{P2BIL}. This later operator randomly mixes two building blocks during an interbreeding, while the cg-combination carefully chooses from which parent to take each position of the probability vector, as described in Section \ref{interbreeding}. In this experiment, the algorithm $\varphi$-PBIL was ran with its default parameters. Additionally, with $k=4$ clusters and population sizes are $N_0=N_w=100$.

Before a quantitative comparison, the typical behavior of both operators is illustrated for a single run for the concatenated trap-5 problem with size 15. The problem is decomposable, and for each subproblem there is an optimal building block. The three optimal building blocks -- BB0, BB1 and BB2 -- must be found and subsequently combined in order to reach the global solution. 

Finding all optimal building blocks is not difficult when the PV-wise uniform crossover is used, as illustrated in figure \ref{bad_clusters}; the difficult thing is to mix them. It is noticeable that each cluster specializes in a specific pattern. Cluster 0, for instance, started attracting some individuals possessing BB2. As the evolutionary process continues, the proportion of individuals possessing BB2 which are attracted to cluster 0 grows, until around 500 fitness evaluations, when 100\% of the individuals of cluster 0 possess BB2. Similarly, cluster 2 specializes on BB0 and cluster 3 on BB1. A different setting may occur if another run is performed for a different random seed, but the general behavior is similar. The global solution, however, was not found.

Running $\varphi$-PBIL with the cg-combination operator for the same random seed (therefore the same initial population) leads to a much better behavior, as figure \ref{good_clusters} shows. In the first steps of the evolutionary process (roughly until 400 fitness evaluations) speciation is clear. Subsequently, as the number of successful combinations grows, clusters attract individuals possessing two or even three optimal building blocks simultaneously. Cluster $3$, finally, specializes on the global optima of the problem after around $2,000$ fitness evaluations. At that moment the algorithm converges, since all positions of all PVs saturate above $0.95$ or below $0.05$. It is interesting to note that some individuals possessing only one correct building block are still in the population even after the combination process is at advanced stages.

A more quantitative comparison between both interbreeding operators is shown in table 2.
Four representative problems were chosen: shuffled HIFF, concatenated trap-5, overlapping concatenated trap-5 and twomax, with problem sizes 128, 100, 60 and 100 respectively. Table 2
reports the fraction of success (Succ. \%), which is the fraction of runs where at least one global optima is found, and statistics for the number of fitness evaluation (Eval.) until convergence of $\varphi$-PBIL.

Using the cg-combination, at least one global optimum is found in 97\% of the runs for HIFF and in 100\% of the runs for the other problems. Actually, when this operator is used, all global optima were found in 93\% of the runs for HIFF and in 100\% of the runs for the other problems. The PV-wise uniform crossover was not able to reach any global optima for any of the problems tested, except for twomax, which is a very simple problem without a complex structure. The results are not surprising as it is well known that blind crossover operators that don't respect the structure of the problem are condemned to fail on these types of problems. A very similar behavior is observed when no interbreeding is performed, but the results are not reported in the table.





\subsection*{Multimodal optimization and symmetry}

\begin{table}[t]
\begin{center}
\begin{small}
\caption{ Effectiveness and efficiency of $\varphi$-PBIL and UEBNA for 10 instances of the graph bisection problem as the mean $\pm$ standard deviation of the number of global optima (peaks) found and number of fitness evaluations after 10 independent runs of each algorithm for each problem. Data for UEBNA extracted from \cite{pena2005}.
}
\begin{tabular}[h]{|l|l|r@{$\pm$}lr@{$\pm$}l|} 
\hline
Problem & \hspace{27pt}EDA & Optima&sd&Eval.&sd \\
\hline
Pgrid16 & UEBNA $k=4$ & 2.0&0.0 & 51400 & 2366 \\
\cline{2-6}
(2 peaks) &$\varphi$-PBIL $k=5$ & 2.0 & 0.0 & 10126 & 606 \\ \hline
Pgrid36 & UEBNA $k=2$ & 2.0&0.0 & 85600 & 8462 \\
\cline{2-6}
(2 peaks) &$\varphi$-PBIL $k=5$ & 2.0 & 0.0 & 28963 & 10754 \\ \hline
Pgrid64 & UEBNA $k=4$ & 2.0&0.0 & 124900 & 3479 \\
\cline{2-6}
(2 peaks) &$\varphi$-PBIL $k=10$ & 2.0 & 0.0 & 64245 & 10999 \\ \hline
Pcat28 & UEBNA $k=2$ & 2.0&0.0 & 57100 & 2846 \\
\cline{2-6}
(2 peaks) &$\varphi$-PBIL $k=5$ & 2.0 & 0.0 & 14311 & 1299 \\  \hline
Pcat42 & UEBNA $k=2$ & 2.0&0.0 & 73900 & 1449 \\ 
\cline{2-6}
(2 peaks) &$\varphi$-PBIL $k=10$ & 2.0 & 0.0 & 29714 & 3644 \\ \hline
Pcat56  &UEBNA $k=4$ & 2.0&0.0 & 96400 & 2366 \\ 
\cline{2-6}
(2 peaks) &$\varphi$-PBIL $k=10$ & 2.0 & 0.0 & 46151 & 4362\\  \hline
Pcatring28 & UEBNA $k=2$ & 4.0&0.0 & 54700 & 949 \\ 
\cline{2-6}
(4 peaks) &$\varphi$-PBIL $k=5$ & 4.0 & 0.0 & 12694 & 853  \\  \hline
Pcatring56 & UEBNA $k=8$ & 3.8&0.4 & 96400 & 1897\\ 
\cline{2-6}
(4 peaks)&$\varphi$-PBIL $k=10$ & 3.9 & 0.3 & 48837 & 12115   \\  \hline
Pcatring42 & UEBNA $k=6$ & 5.9&0.3 & 75700 & 3302\\ 
\cline{2-6}
(6 peaks)&$\varphi$-PBIL $k=15$ & 6.0 & 0.0 & 32361 & 1513    \\  \hline
Pcatring84 & UEBNA $k=10$ & 4.8&0.8 & 121000 & 3162 \\ 
\cline{2-6}
(6 peaks)&$\varphi$-PBIL $k=20$ & 5.7 & 0.7 & 84539 & 9300  \\  \hline

 \end{tabular}
 \end{small}
\end{center}
\label{tab1}
\end{table}

This experiment validates $\varphi$-PBIL on multimodal optimization problems, some of which present interactions among genes. Those features require good niching and linkage learning capabilities from the evolutionary algorithm. A through comparison between $\varphi$-PBIL and UEBNA is described. The latter was already shown to be competent to solve globally multimodal optimization problems. In \cite{pena2005} UEBNA is compared to other EDAs when solving of instances globally multimodal problems. Experiments reported in that work clearly show that UEBNA achieves much better efficiency and efficacy when compared to EBNA, which is based on the supervised induction of Bayesian networks and does not incorporate cluster labels.

The graph partitioning problem was used in this comparison. 
Some instances of this problem are hard for EDAs. Particularly, larger ``ring'' topologies (Pcatring56, Pcatring42 and Pcatring84) are challenging because they clearly present a structure of interactions among genes, besides multimodality and symmetry. In a binary codification of one gene per node there are small groups of 7 nodes which should be identified and combined until all global solutions are found. Figure \ref{topologia} illustrates some of those instances.

All results for UEBNA presented here were extracted from \cite{pena2005}. Ten independent runs of each algorithm are performed for each problem instance. A fixed set of parameters for $\varphi$-PBIL is used for all instances: $p_w=75\%$, $p_{old}=25\%¨$ and $p_c=50\%$ (using all $50\%$, which are the default parameters, resulted slightly inferior results). The initial population size $N_0=4,000$ is the same for $\varphi$-PBIL and UEBNA. Additionally, the working population size of $\varphi$-PBIL is set as $N_w=500$. The only free parameter adopted here is $k$, the number of clusters. The value of $k$ which empirically maximizes the performance of the algorithm is chosen, for $\varphi$-PBIL, from the interval $k \in [2,3,\ldots,20]$ and, for UEBNA, from the results reported in \cite{pena2005}. A similar comparison to UEBNA was already reported in \cite{EP07a}, but $\varphi$-PBIL was run with default parameter set and the repair operator for the problem of graph partitioning was not used there.

Table 3 summarizes the results. The problems and algorithms are organized in rows and the data columns show mean $\pm$ standard deviation of the number of global optima found (\textit{Optima}$\pm$\textit{s.d.}) and for mean $\pm$ standard deviation (\textit{Eval.}$\pm$\textit{s.d.}) of the number of fitness evaluations performed until convergence, evaluated over 10 runs. All runs of both algorithms found at least one global optima for all instances tested.

It is clear from results that $\varphi$-PBIL attains convergence for all global optima using less fitness evaluations. For smaller problem instances both algorithms find all global optima in all runs but, for greater problem instances, $\varphi$-PBIL presents better efficiency. The mean number of global optima found is better than for UEBNA, noticeably for Pcatring56, Pcatring42 and Pcatring84. For the later, which presents $6$ global optima,  $\varphi$-PBIL found $5.7\pm0.7$ global optima, while UEBNA found $4.8\pm0.8$ global optima. This may suggest that $\varphi$-PBIL presents a good scalability behavior. Further scalability tests were also shown in \cite{EP07a}.

$\varphi$-PBIL can attain good results with relatively small populations. It is also important to note that computational complexity of the learning step of $\varphi$-PBIL is related to a k-means update, what is less time-consuming than learning Bayesian networks. Howerver, when comparing computational complexity of $\varphi$-PBIL to other algorthms one should consider that the model is is updated after the introduction of a single individual, while most algorithms learn a model for each new population.

\section{Conclusion}

This work presents $\varphi$-PBIL: an evolutionary algorithm based on the application of an interbreeding operator, which is guided by the information obtained by clustering the population of an EDA.  Identification of substructures is performed by clustering and combination among substructures can be achieved by combining information from different clusters. The core algorithm is a simple clustered EDA that adopts order-2 probabilistic models, which leads to a very parsimonious approach when compared to other competent EDAs. Its simple incremental approach is also noticeable, where individuals are successively generated from continuously updated binomial probabilistic models.

Probabilistic models used by $\varphi$-PBIL are much simpler to infer when compared to Bayesian networks, which are currently the most effective models for EDAs. Actually, $\varphi$-PBIL keeps probabilistic models updated after each new individual is generated and selected at the cost of a single k-means update.

The most sophisticated part of the algorithm is the interbreeding mechanism, which respects the problem structure and relies on the diversity of those substructures as one of the causes for the existence of clusters in the population. Experiments show some evidence that this mechanism is very important for solving problems that present interactions among the variables. Literature generally recommends avoiding the combination from different clusters, since this is often unproductive. $\varphi$-PBIL attempts to minimize bad results by carefully choosing how to combine the most informative portions.

An empirical search for default values for some parameters was performed, but a more formal investigation should also be done. Furthermore, it is important to understand how to set the population size and number of clusters for each kind and size of problem. Some work on this topic has already been done for other evolutionary algorithms \cite{lobo99} and the methodology could be adapted for our algorithm. The relation between some of the parameters should also be better explored. For instance, the number $k$ of clusters should be related to the population size $N_w$. Actually, some empirical investigation (not shown) has pointed out that $N_w/k$ should be of the order of magnitude of $10^1$. Actually, $N_w/k= 20$ or $30$ was enough for most problems verified.

Future work will check for relevant features, like scalability, for a wider range of problems including HDFs like HIFF \cite{hiff}. Further empirical investigation on the behavior of the interbreeding mechanism should also be performed in order to better explain how exploration is actually done for several classes of problems. 

\begin{section}{Acknowledgements}
\nonumber
The authors would like to thank Fernando Lobo for his valuable contribution to improve this paper.
\end{section}



%


\end{document}